\documentclass{article}

\usepackage[nonatbib, preprint]{neurips_2019}

\usepackage[utf8]{inputenc} 
\usepackage[T1]{fontenc}    
\usepackage{hyperref}       
\usepackage{url}            
\usepackage{booktabs}       
\usepackage{amsfonts}       
\usepackage{amsmath}
\usepackage{nicefrac}       
\usepackage{microtype}      
\usepackage{graphicx}
\usepackage{booktabs}
\usepackage{multirow}
\usepackage{siunitx}

\usepackage{algorithm}
\usepackage{algorithmic}

\renewcommand{\arraystretch}{1.2}

\title{Density estimation in representation space to predict model uncertainty}

\author{%
  Tiago Ramalho \\
  Cogent Labs\\
  Tokyo, Japan\\
  \texttt{tramalho@cogent.co.jp} \\
   \And
   Miguel Miranda \\
   Apple Inc. \\
   Cupertino, United States\\
   \texttt{miguelnmiranda@apple.com} \\
}

\expandafter\def\expandafter\normalsize\expandafter{%
    \normalsize
    \setlength\abovedisplayskip{4pt}
    \setlength\belowdisplayskip{4pt}
    \setlength\abovedisplayshortskip{4pt}
    \setlength\belowdisplayshortskip{4pt}
}

\begin{document}

\maketitle

\begin{abstract}
Deep learning models frequently make incorrect predictions with high confidence when presented with test examples that are not well represented in their training dataset.
We propose a novel and straightforward approach to estimate prediction uncertainty in a pre-trained neural network model. 
Our method estimates the training data density in representation space for a novel input. A neural network model then uses this information to determine whether we expect the pre-trained model to make a correct prediction.
This uncertainty model is trained by predicting in-distribution errors, but can detect out-of-distribution data without having seen any such example.
We test our method for a state-of-the art image classification model in the settings of both in-distribution uncertainty estimation as well as out-of-distribution detection.
\end{abstract}

\section{Introduction}
\label{sec:intro}
Deep learning methods have delivered state-of-the art accuracy in challenging tasks such as image classification, language translation, voice-to-speech or learning to act in complex environments. Yet these methods can make incorrect confident predictions when shown certain data~\cite{nguyen_deep_2014,amodei_concrete_2016}. 
As an example consider a deep convolutional network~\cite{krizhevsky2012imagenet} for image classification trained with a fixed set of target labels. When shown an image with a class label missing from the original set, the model will (often confidently) predict one of the classes in the training set even though none of them are correct~\cite{hendrycks_baseline_2016}.

At the heart of this issue is the fact that we train models to map a certain data probability distribution to a fixed set of classes. In the real world these models must cope with inputs which fall out of the training distribution. This can be either due to adversarial manipulation~\cite{athalye_synthesizing_2017,huang_adversarial_2017}, drifts in the data distribution or unknown classes. 

This kind of uncertainty in the inference result, also known as epistemic uncertainty, cannot be reduced even as the size of the training set increases~\cite{sensoy2018evidential,osband_randomized_2018}. To address epistemic uncertainty it is therefore necessary to add a prior to the model which allows for the existence of out-of-distribution data.

In this work we aim to add such a prior by estimating the distance in representation space between a new test point and its closest neighbors in the training set. If the model under consideration is well fit to the training dataset, we expect that distance in a high level representation space will correlate well with distance in the distributional sense. In the following section, we test this hypothesis empirically for the ImageNet dataset and verify that it holds true.

Based on this hypothesis, we propose training a neural network model which takes as inputs the statistics of a point's neighborhood in representation space, and outputs the probability that the original inference model will make a mistake. This model can be trained using only the original model's training dataset and does not need to train on a separate validation or out-of-distribution dataset. 

We compare this method with naive density estimates using a point's nearest neighbors in representation space, as well as with baseline methods such as the predictive uncertainty after the softmax layer and a method based on building gaussian estimates of each class' density. Our method outperforms these baselines and shows a way forward to integrate uncertainty estimates into existing deep neural network models with minor architectural modifications.

In short, the main contributions of this paper are:
 
\begin{itemize}
   \item We show that a neural network classification model trained on Imagenet has higher accuracy when the representation of a new data point is close to previously seen representations of that same class. This holds true even for incorrectly classified examples in the training set.
   \item We use the above observation to propose a simple method to detect out-of-distribution data as well as misclassifications for existing trained models. Crucially, this method does not need to be trained with an external out-of-distribution dataset. 
    \item We test the performance of our method in the challenging scenario of Imagenet image classification and show that it outperforms several baselines. Out-of-distribution prediction is tested on several datasets including Imagenet classes outside of the ILSCVR2012 set, Imagenet-C, and Imagenet-V2.
\end{itemize}

\section{Representation space in classification models}

Consider a dataset with $N$ labeled pairs $x_i, y_i$ with $x_i$ usually a high dimensional data element and $y_i$ a lower dimensional label. We wish to train a model to minimize the error in the estimator $\text{argmax}_y P_{\theta}(y|x)$, with $P_{\theta}$ the probability distribution for the labels output by the model. 

This is done by minimizing the cross-entropy between $P_{\theta}$ and the true distribution $P$ of \textit{the training data}. This loss is minimized for points drawn from $P_{\text{train}}(x)$, the training data distribution. We expect a model to generalize well when the support of $P_{\text{test}}(x)$ matches that of $P_{\text{train}}(x)$.

\begin{figure}[t]
  \centering
    \includegraphics[width=0.35\textwidth]{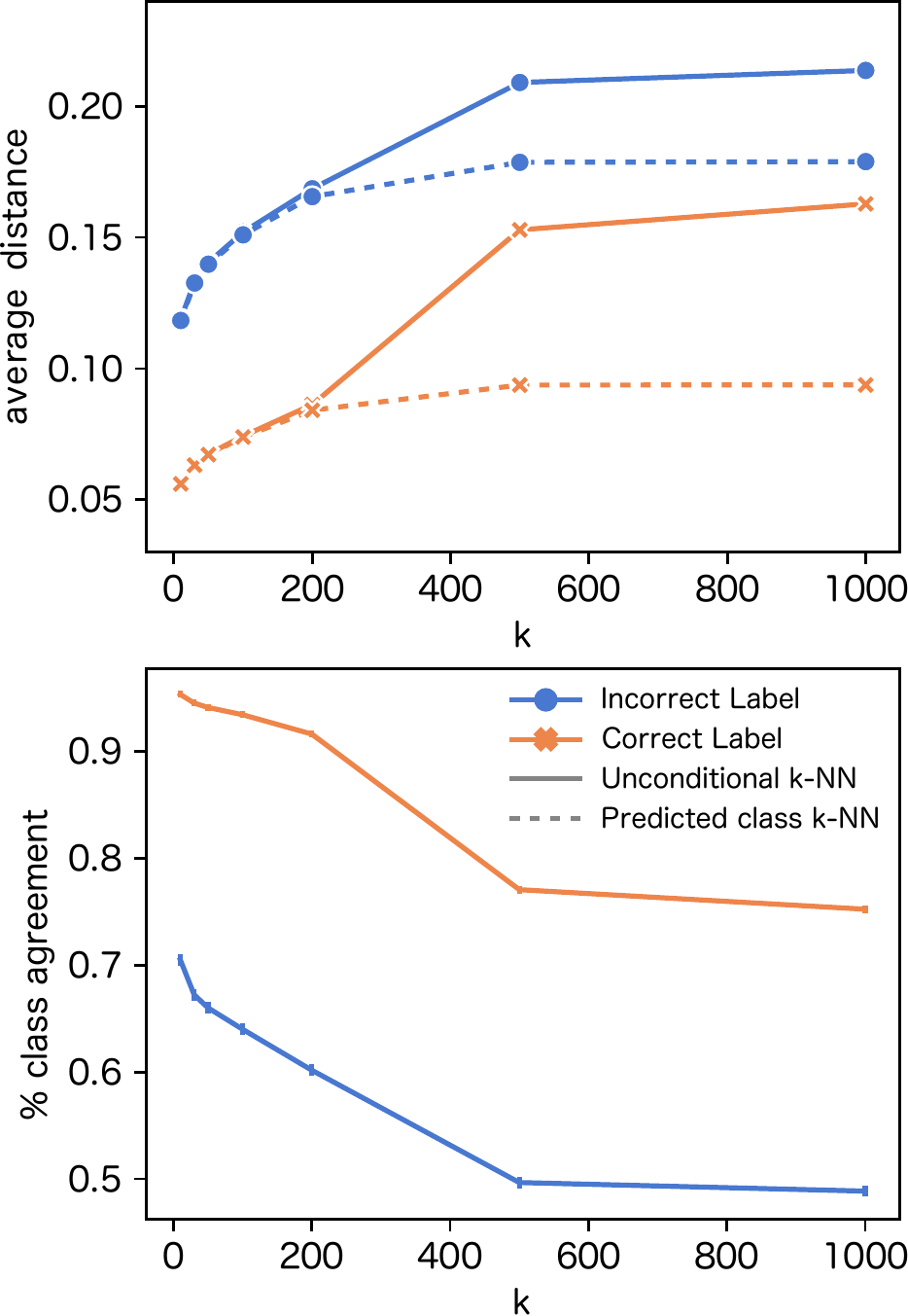}
    \caption{Quantifying the neighborhood statistics (ILSVRC2012 validation set) of the representation space of the last layer for Inception-ResNet-v2~\cite{szegedy_inception-v4_2016}. Top, we plot the average distance Eq. \ref{eq:kde} (solid), \ref{eq:kdecond} (dashed) between a new representation and its nearest neighbors in the training set conditioned on whether that point was correctly or incorrectly classified. Bottom, class label agreement Eq. \ref{eq:agreement} as a function of $k$. Incorrectly classified data are both further away from their closest neighbors and their predicted labels disagree more with the target prediction.} 
    \label{fig:statistics}
\end{figure}

\begin{figure*}[t!]
  \centering
    \includegraphics[width=0.6\textwidth]{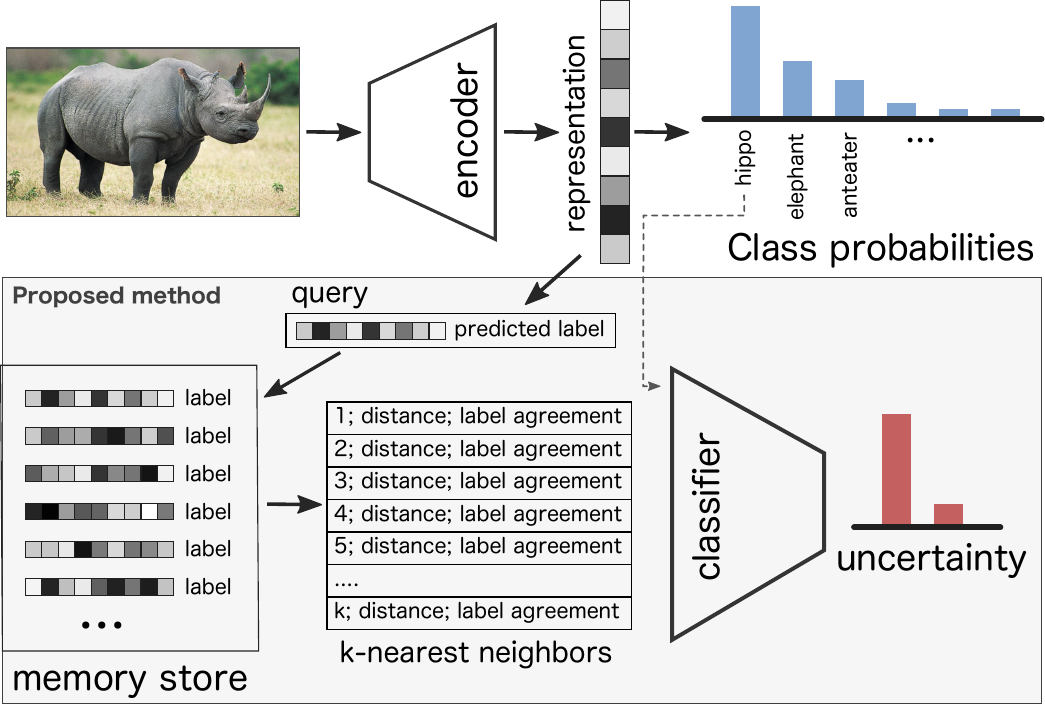}
    \caption{Schematic of the proposed model for uncertainty estimation. Top row, a deep neural network model is trained using the standard cross-entropy loss to produce class probabilities within a known set. The high level representation (before the logits layer) is additionally stored in a queriable memory database. When a new input is classified, its representation can be used to query the database, and information about its nearest neighbors is fed to the uncertainty model, which predicts the likelihood of the classification result being incorrect.}
    \label{fig:model}
\end{figure*}

Most interesting datasets are very high dimensional, which makes a direct approximation of the training distribution density computationally challenging. To alleviate this problem we propose performing the estimation procedure in the high-level representation space used by a neural network model, which can be seen as a compressed representation of the data.

A typical deep neural network model with $N$ layers will transform $x$ via a series of nonlinear maps $r_i = f^i(r_{i-1})$ (with $r_0 = x$). A linear mapping is then applied to the final representation $r$ to obtain the logits $\ell = f^N(r_{N-1})$, which when passed through the softmax function are taken to be an estimate of the class conditional probability distribution $P(y|x)$.

These representations $r_i$ are lower dimensional than $x$ and in particular the last representation $r_N$ in a trained model will have a high mutual information with the variable $y$ (as the posterior $P(y|x)$ is just a linear function of $r_N$). Therefore, in this space we expect data points with similar classes to roughly cluster together which makes it feasible to use k-nearest neighbors to calculate an approximation of the training set density.

To verify this assumption, we can calculate several summary statistics on the representation space generated by applying a fully trained Inception-ResNet-v2 model~\cite{szegedy_inception-v4_2016} to the ImageNet ILSVRC2012 dataset~\cite{ILSVRC15}.

Let $r_i$ be the representation of input $x_i$; $y_i$ be its corresponding ground truth label and $\hat{y}_i$ the model's prediction for the class label. Define $ \mathcal{N}(r_i, k)$ as set of k-nearest neighbors of $r_i$. Then, we define the following statistics calculated on the k-nearest neighbors of point $x_i$:

The unconditional kernel density estimate of $x_i$ with representation $r_i$ 
\begin{equation}
    P(x_i)\propto \sum_{j\in\mathcal{N}(r_i)}^k d(r_i, r_j)    
    \label{eq:kde}
\end{equation}

with $d$ a suitable kernel (such as the $\ell_2$ norm or the cosine similarity).

The class conditional kernel density estimate, which considers only neighbors with the same class as that predicted by the model:
\begin{equation}
    P(x_i|\hat{y}_i)\propto \sum_{j\in\mathcal{N}(r_i):\, y_j = \hat{y}_i}^k d(r_i, r_j)
    \label{eq:kdecond}
\end{equation}

And  the binary agreement density estimate of $x_i$, which does not use distance information in representation space, but instead counts how many neighbors match the predicted label $\hat{y}$:
\begin{equation}
    P(x_i|\hat{y}_i) \propto  \sum_{j\in\mathcal{N}(r_i):\, y_j = \hat{y}_i}^k \mathbb{I}\left(y_j=\hat{y}_i\right)
    \label{eq:agreement}
\end{equation}

This is equivalent to calculating an unweighted kernel density estimate in a ball of radius $\epsilon$ with $\epsilon = \text{max}\left( d(r_i, r_j)\right)$. While this is a high variance estimate of $P(x|y)$, it is computationally efficient especially if we are computing an estimate of Eq.~\ref{eq:kde} at the same time.

We plot the three statistics defined above as a function of $k$ in Image \ref{fig:statistics}. We use as test points the images in the validation set, and query the representations of the whole training set to compute the statistics. As hypothesized, points where the network makes a misclassification are significantly more distant than their nearest neighbors, and their neighbors' ground truth labels tend to disagree more with the network's predicted label. The distances for the class-conditional KDE grow more slowly as the effective number of neighbors gets smaller with increasing $k$.

\section{Method}
\label{sec:classifier}

Motivated by the observations above, we wish to propose a method which can use this information to determine whether a new test point is likely to be close to the training data distribution. While the above statistics could be used directly, the density of the training set in the representation space is unlikely to be linearly correlated to the final accuracy of the model. Indeed, in the results section we show that these statistics used directly as an uncertainty measure do not offer as good performance as the model proposed below.

Formally we are looking for a model which can output an uncertainty score given the neighborhood information:
\begin{equation}
    u(x_i) = g_{\theta}(\{(r_j, \mathbb{I}\left(y_j=\hat{y}_i\right)\}_{j \in \mathcal{N}(r_i)}, s(\hat{y}_i))
    \label{eq:classifier}
\end{equation} 

with $s(\hat{y}_i)$ the original model's posterior probability of the predicted label $\hat{y}_i$.

We propose implementing $g_{\theta}$ as a feedforward neural network in order to be able to capture any nonlinear interactions. Furthermore, we want to learn the parameters $\theta$ of this model without resorting to any out-of-distribution information. To achieve this we will rely on the related task of detecting a model's mistakes to train the model. This task only requires in-distribution data where the model still makes some mistakes.

We can therefore use either a subset of the validation set or the training set itself as long as the model has not been trained to reach 0\% training error. This assumption is often true on state-of-the art models as they are trained with some form of regularization (weight decay, early stopping, etc...). In this case, the mistakes will be data points that are harder to fit as they are more likely to be outliers far from the majority of data.

In this dataset we minimize the binary cross entropy loss \begin{equation}
\mathcal{L}[u, t] = -t_i \log\left(u(x_i)\right)
\label{eq:loss}
\end{equation}
with the binary labels $t_i=1$ if $y_i = \hat{y}_i$ and $t_i=0$ otherwise.

As the order with which the model is presented with its inputs should not vary, we use an architecture invariant to permutations in the inputs via the use of an aggregation step   \cite{zaheer_deep_2017,garnelo_conditional_2018}. $g_\theta$ is a model with $L$ layers, where each hidden layer is of the form

\begin{equation}
    h^l_i = \sum^k_{j\in\mathcal{N}(r_i)} q^l(h_j^{l-1})
    \label{eq:model}
\end{equation}

Where $q$ is a linear transform. For $l > 1$, a nonlinearity is applied to the hidden representation after the aggregation. In this case $L=1$ corresponds to a linear model. At the very last layer $l=L$ a linear layer with output dimensionality 2 is applied to the aggregated representation to obtain the classification logits.

For the remainder of this paper we will adress our proposed method as Neighborhood Uncertainty Classifier (\textbf{NUC}). The full training procedure is summarized in Algorithm \ref{alg:classifier}. The algorithm requires a trained model $f$ which returns unnormalized log-probabilities for $\hat{y}$; a precomputed index of all representations at the final layer which we denote by $\mathcal{A}$ for all points in the training set; and the choice of hyperparameter $k$, the number of nearest neighbors. 

For every point in the dataset we obtain the representation and query its $k$ nearest neighbors (excluding the exact point, which will be in the index too). This gives us the set of all neighbor representations as well as their ground truth labels $\{(r_j, y_j)\}$. This information is fed to the model $g$ as well as the predicted class's confidence $s(\hat{y}_i)$ and used to perform one step of gradient descent.

\begin{algorithm}
\caption{NUC training procedure}
\begin{algorithmic}[1]
\REQUIRE{$k$ number of neighbors, $\mathcal{A}$ the set of all representations, $\theta$ model parameters, $f$ a trained model}
\FOR{epoch $\in$ number of epochs}
    \FOR{$x_i \in$ training set}
        \STATE $r_i \gets \text{f}^{N-1}(x_i)$
        \STATE $\hat{y}_i \gets \text{argmax} \; \text{f}^N(r_i)$
        \STATE $t \gets (y_i = \hat{y}_i)$
        \STATE $\{(r_j, y_j)\}_{j \in \mathcal{N}(r_i)} \gets \text{knn}(r_i, k, A)$
        
        \STATE $u(x_i) \gets g_{\theta}(\{(r_j, \mathbb{I}\left(y_j=\hat{y}_i\right)\}_{j \in \mathcal{N}(r_i)}, s(\hat{y}_i))$
        \STATE $\theta_{t+1} \gets \theta_t + \lambda \nabla_{\theta} \mathcal{L}\left[u(x_i), t\right]$
    \ENDFOR
\ENDFOR
\end{algorithmic}
\label{alg:classifier}
\end{algorithm}

\section{Experiments}

\begin{table*}[t]
\begin{center}
\footnotesize
\renewcommand{\arraystretch}{1.5}
\begin{tabular}{ lccc } 
 & ILSVRC2012$_{\text{valid.}}$ & Imagenet-V2 \cite{recht2019imagenet} & Imagenet-C \cite{hendrycks_benchmarking_2019} \\
\cline{2-4}
 & \multicolumn{3}{c}{AUROC / AUPR-Out / AUPR-In} \\
\hline
Softmax & 0.844 / 0.587 / 0.945 & 0.826 / 0.682 / 0.896 & 0.848 / 0.812 / 0.852 \\
Softmax$^\dagger$ & 0.848 / 0.590 / 0.948 & 0.829 / 0.681 / 0.899 & 0.849 / 0.811 / 0.854 \\
Eq.~\ref{eq:kde} & 0.772 / 0.389 / 0.930 & 0.781 / 0.525 / 0.893 & 0.826 / 0.730 / 0.862 \\
Eq.~\ref{eq:kdecond} & 0.773 / 0.398 / 0.931 & 0.781 / 0.541 / 0.892 & 0.829 / 0.743 / 0.865 \\
Eq.~\ref{eq:agreement} & 0.818 / 0.470 / 0.950 & 0.815 / 0.602 / 0.914 & 0.844 / 0.759 / 0.883 \\
Mahalanobis & 0.786 / 0.462 / 0.931 & 0.798 / 0.608 / 0.896 & 0.842 / 0.789 / 0.869 \\
NUC & \textbf{0.862} / \textbf{0.600} / \textbf{
0.959} & \textbf{0.845} / \textbf{0.690} / \textbf{0.923} & \textbf{0.862} / \textbf{0.820} / \textbf{0.889} \\
\end{tabular}
\end{center}
 \caption{Results for the in-distribution uncertainty quantification task. We predict classification mistakes (top-1) for the ILSVRC2012 validation set, Imagenet-V2 \cite{recht2019imagenet} new validation set and the Imagenet-C \cite{hendrycks_benchmarking_2019} dataset. We report the threshold independent metrics AUROC, AUPR-Out and AUPR-In \cite{hendrycks_baseline_2016}. Items marked with $^\dagger$ used the calibration procedure described in~\cite{guo_calibration_2017}.} 
 \label{tab:ind}
\end{table*}

We consider two scenarios where the proposed method could be useful: in-distribution uncertainty estimation (i.e. predict incorrect classifications); and out-of-distribution detection. We wish to perform these tests on a real world dataset of practical use, and we therefore choose the widely used Imagenet ILSVRC2012 set. Unlike datasets such as MNIST or CIFAR-10 where performance is saturated~\cite{lee_simple_2018} and any differences between methods are difficult to see, ILSVRC2012 is challenging enough to demonstrate the difference between different performing methods.

To generate representations, we use a pretrained checkpoint for the Inception-ResNet-v2 network~\cite{szegedy_inception-v4_2016} as it comes close to state of the art accuracy on this dataset and is very widely accessible. The 1.2 million 1536-dimensional vectors needed to represent the whole training set require 5GB of memory which allows us to keep the complete index in memory in a modern workstation.

We then train the model described in Section \ref{sec:classifier} with $L=2$ by following the procedure in Algorithm \ref{alg:classifier} using the Adam optimizer with a learning rate of $10^{-3}$ annealed to $10^{-4}$ after 40000 steps. We train the model for one single epoch before testing.

We calculate the performance of all models by a series of threshold-independent measures: the AUROC, AUPR-In and AUPR-Out, as suggested in \cite{hendrycks_baseline_2016}. For NUC, we take the softmax confidence that the model will make a mistake as the input to the threshold-independent metrics.

\begin{figure}[t]
  \centering
    \includegraphics[width=0.45\textwidth]{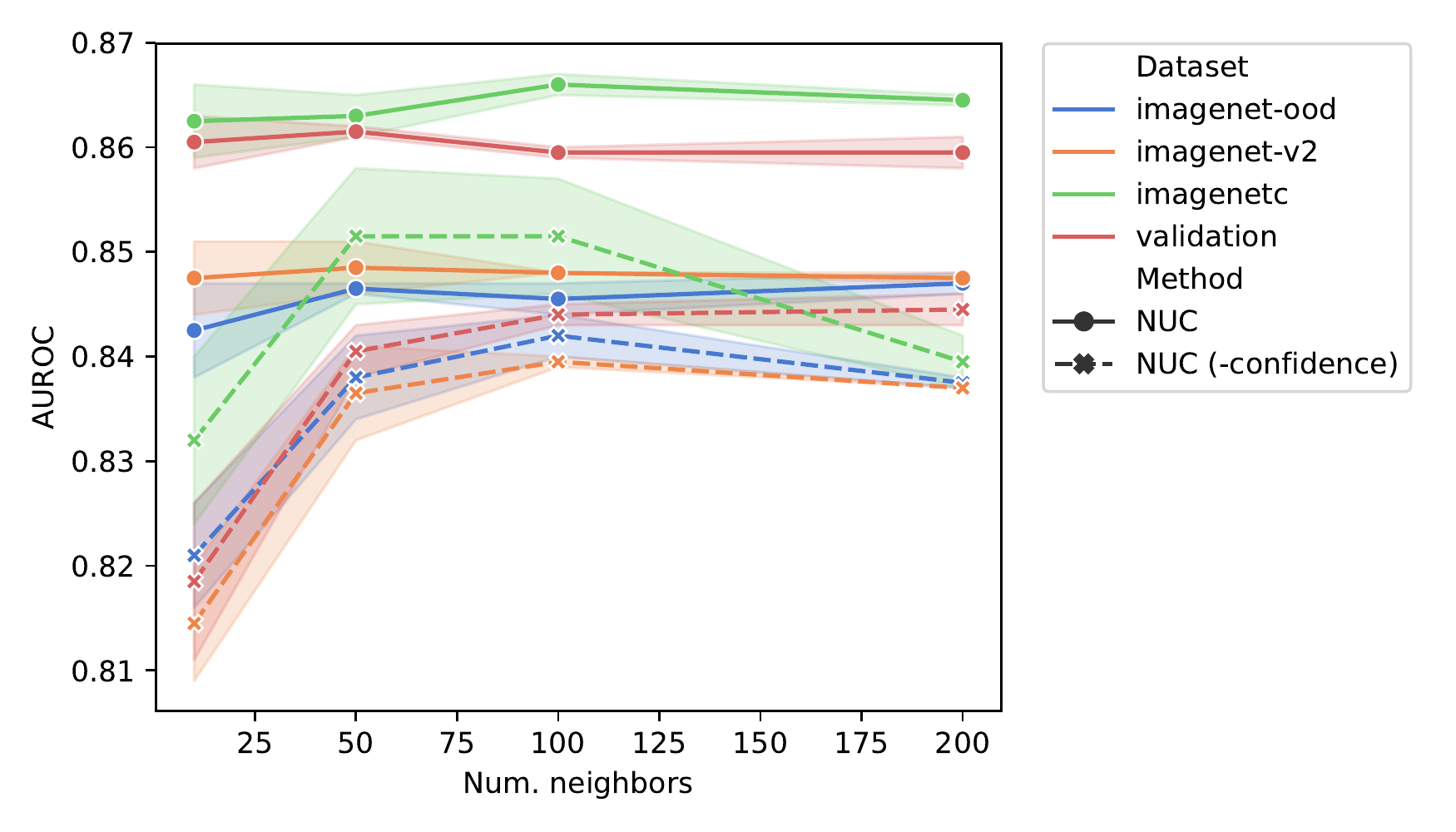}
    \caption{NUC performance as a function of number of neighbors $k$. Without taking the softmax prediction confidence into account, the model's performance is strongly dependent on $k$, as expected. For large values of $k$ the performance slightly degrades as neighbors from potentially very far regions are being taken into account. The full model's performance, however, is relatively constant throughout different values of $k$.}
    \label{fig:comparison}
\end{figure}

In Figure \ref{fig:comparison} we plot NUC's AUROC as a function of the hyperparameter $k$. We observe that the method's performance is relatively constant for most values of $k$. If NUC does not have access to the original model's confidence, however, the performance becomes strongly sensitive on $k$. This would be expected, as in this case all the inputs of the model depend on the neighborhood information. We can even observe a degradation in performance for very large values of $k$, as the neighborhood grows more sparse and neighbors from farther away regions are collected. This is consistent with the observations in Figure \ref{fig:statistics}.

This issue is fixed in the full model, which shows better and more stable performance as it has access to two different sources of information. Detailed dataset comparisons are provided in the following section. For the remainder of our analysis we set $k=10$.

\subsection{In-distribution uncertainty detection results}

For this task, we use datasets with images following the same distribution as that of the original ILSVRC2012 training set: the ILSVRC2012 validation set; Imagenet-V2 \cite{recht2019imagenet}, an alternative validation set with subtle distributional shifts (we use the \textit{MatchedFrequency} subset as it is the most challenging); and the Imagenet-C \cite{hendrycks_benchmarking_2019} dataset which adds a number of visual perturbations to the images which leave the class unchanged to a human observer.

We use the following methods as comparative baselines: the softmax confidence (both original model's and calibrated confidence on the validation set as in ~\cite{guo_calibration_2017}); the kernel density estimates Eq.~\ref{eq:kde} (average distances), Eq.~\ref{eq:agreement} (neighborhood agreement), Eq~\ref{eq:kdecond} (conditional distances) using 200 $k$-nearest neighbors; and the mahalanobis distance to the predicted class's centroid (calculated following~\cite{lee_simple_2018}).

We report results in Table \ref{tab:ind}. In the case of misclassification prediction for the Imagenet validation dataset NUC beats all the baselines. We note that pure distance based methods don't fare as well as the softmax confidence, possibly because the test points are still close to the training set and misclassifications come from the model not being able to distinguish the data rather than data points finding themselves in unusual places in representation space.

 Our method also outperforms other baselines for the Imagenet-V2 and Imagenet-C datasets, which can be considered in-distribtion (as they are images containing the same classes as found on the original ILSVRC2012 class set).

\subsection{Out-of-distribution detection results}

\begin{table*}[t]
\begin{center}
\footnotesize
\renewcommand{\arraystretch}{1.5}
\begin{tabular}{ lccc } 
 & Imagenet$_{\text{unk.}}$ & SVHN \cite{svhn_2011} & CIFAR-10 \\
\cline{2-4}
 & \multicolumn{3}{c}{AUROC / AUPR-Out / AUPR-In} \\
\hline
Softmax & 0.818 / 0.832 / 0.767 & 0.938 / 0.948 / 0.912 & 0.974 / \textbf{0.984} / 0.945  \\
Softmax$^\dagger$ & 0.821 / 0.831 / 0.773 & 0.938 / 0.948 / 0.916 & 0.975 / \textbf{0.984} / 0.947  \\
Eq.~\ref{eq:kde} & 0.816 / 0.767 / 0.812 & 0.932 / 0.896 / 0.922 & 0.952 / 0.863 / 0.924  \\
Eq.~\ref{eq:kdecond} & 0.816 / 0.771 / 0.812 & 0.932 / 0.899 / 0.923 & 0.946 / 0.838 / 0.922  \\
Eq.~\ref{eq:agreement} & 0.822 / 0.786 / 0.841 & 0.927 / 0.890 / 0.931 & 0.943 / 0.794 / 0.938  \\
Mahalanobis & 0.835 / 0.833 / 0.817 & 0.906 / 0.948 / 0.814 & 0.908 / 0.950 / 0.813  \\
NUC & \textbf{0.846} / \textbf{0.846} / \textbf{0.830} & \textbf{0.948} / \textbf{0.953} / \textbf{0.942} & \textbf{0.976} / 0.979 / \textbf{0.958} \\
\end{tabular}
\end{center}
 \caption{Results for the out-of-distribution detection task. The in-distribution set is composed of correctly classified ILSVRC2012 validation set images, and as out-of-distribution sets we consider the following: Imagenet images with unknown classes (details in main text); SVHN \cite{svhn_2011}; CIFAR-10. We report the threshold independent metrics AUROC, AUPR-Out and AUPR-In \cite{hendrycks_baseline_2016}. Items marked with $^\dagger$ used the calibration procedure described in~\cite{guo_calibration_2017}.} 
 \label{tab:ood}
\end{table*}

To test our model in the out-of distribution case we collected 30000 images from the Imagenet database with classes not represented in the original ILSVRC2012 competition's 1000 classes, which we call Imagenet$_{\text{unk.}}$. All out-of-distribution datasets are combined with the ILSVRC2012 validation set for positive samples. We also test the model against the SVHN and CIFAR-10 datasets. We report the same metrics as before on all the baselines.

For the Imagenet$_{\text{unk.}}$ dataset NUC significantly outperforms all other baselines. We note that in this case all distance based methods perform well, and especially the previously proposed Mahalanobis distance-based method beats both softmax and naive nearest neighbors based approaches. These results strengthen our confidence that representation space distance-based approaches can be a valuable tool to detect uncertainty.

In the case of SVHN and CIFAR-10, the performance of other methods is quite robust and therefore the results are all close. As before, NUC provides the best overall performance.

\section{Related Work}

Existing methods for classification using deep neural networks deliver high accuracy, but suffer from overly-confident outputs and fragility to out-of-distribution data. While ideally we should work towards model architectures which have a built-in concept of uncertainty, here we focus on the more tractable problem of adding uncertainty estimates to the uncertainty unaware model classes currently in widespread use.

The class probabilities output at the softmax layer level can be seen as a baseline for a model's confidence \cite{hendrycks_baseline_2016}. However, \cite{guo_calibration_2017} show that powerful models tend to be over-confident in their predictions. This can be measured by calculating whether the empirical accuracy on the test set matches the accuracy estimate implied by the model's confidence. Adjusting the temperature of the softmax function can be used to improve calibration \cite{liang_enhancing_2017}.

Further approaches to out-of-distribution detection can be broken down into two categories: either the goal is to calibrate a model's posterior predictive distribution so that out-of-distribution data results in a uniform posterior \cite{hafner2018reliable}; or a secondary procedure is developed which produces an uncertainty estimate separate from the model prediction \cite{oberdiek_classification_2018,malinin2018predictive}.

A number of publications focused on steering the network's outputs towards a more uniform distribution in cases where the output is uncertain \cite{lee_training_2017,lakshminarayanan_simple_2017}. The idea is that such methods would produce more calibrated confidence estimates as well as uncertainty estimates indirectly derived from the entropy of the confidence distribution.

It is also possible to train a network to predict how certain the original model is by allowing the network to scale the loss function with a predicted uncertainty value, thereby being penalized less for making mistakes for more challenging data~\cite{devries_learning_2018}. Similarly to our approach, the result is a second model which predicts the uncertainty of the original. In this case the knowledge about uncertainty is implicitly encoded in the auxiliary network's weights rather than in the density of the training data itself.

A number of prior works have used the idea of measuring the distance from a sample to the data manifold to estimate uncertainty. In~\cite{feinman_detecting_2017} a kernel density estimate of the likelihood of the current point is fed as a feature along with a dropout uncertainty estimate calculated (as in \cite{gal_dropout_2015}) to a classifier which can predict adversarial examples reliably. 

A number of recent papers used conformal methods to determine what is the probability of a new point being within the data distribution we have observed before. \cite{hechtlinger_cautious_2018} builds class-conditional uncertainty estimates using a kernel density estimator for $P(x|y)$. In~\cite{papernot_deep_2018} the conformal framework is also used, but this time the neighborhood agreement across multiple layers is calculated and aggregated to compute a p-value. This value is used as a proxy for uncertainty. 

\cite{jiang_trust_2018} uses the ratio of the distance between the closest points of the predicted class and the next closest class as a proxy for uncertainty. \cite{lee_simple_2018} calculates a class conditional gaussian approximation for $P(x|y)$ in representation space.

A generative model trained on the full data distribution can be used to assess the likelihood of the current test point directly. Such methods have been used in adversarial example detection~\cite{uesato_adversarial_2018}. Unfortunately current generative models are not yet ready for epistemic uncertainty estimation \cite{nalisnick_deep_2018}, as they can assign higher likelihoods to out-of-distribution data than to in-distribution. Further research will be necessary to determine how to deploy them in this scenario.

\section{Conclusions}

In this paper we introduced a new method to detect out-of-distribution data shown to an already trained deep learning model. We proposed training a neural network model which takes in as inputs the statistics of a point's neighborhood in representation space, and outputs the probability that the original inference model will make a mistake. This model can be trained using only the original model's training dataset and does not need to train on a separate validation or out-of-distribution dataset.  In this way it can generalize better to out-of-distribution examples, as it only has access to invariant properties of the representation space.

We tested the performance of our method in the challenging scenario of ImageNet image classification and showed that it outperforms several baselines. Both for in-distribution data (ILSCVR2012 set, ImageNet-C, and ImageNet-V2) and out-of-distribution data (ImageNet new classes, CIFAR10, SVHN), NUC beats all baselines under comparison.

An improvement outside of the scope of this work would be to find a minimal set of support neighbors (similar to Prototype Networks \cite{snell_prototypical_2017}) which can be queried instead of the full training set. We expect distance-based methods such as the one here suggested can be further improved and be used a general tool to introduce uncertainty in deep neural networks.


\bibliography{main}
\bibliographystyle{plain}

\end{document}